\documentclass{bmvc2k}
\newcommand{\figteaser}{
\begin{figure*}[t]
\centering
\includegraphics[width=0.9\linewidth]{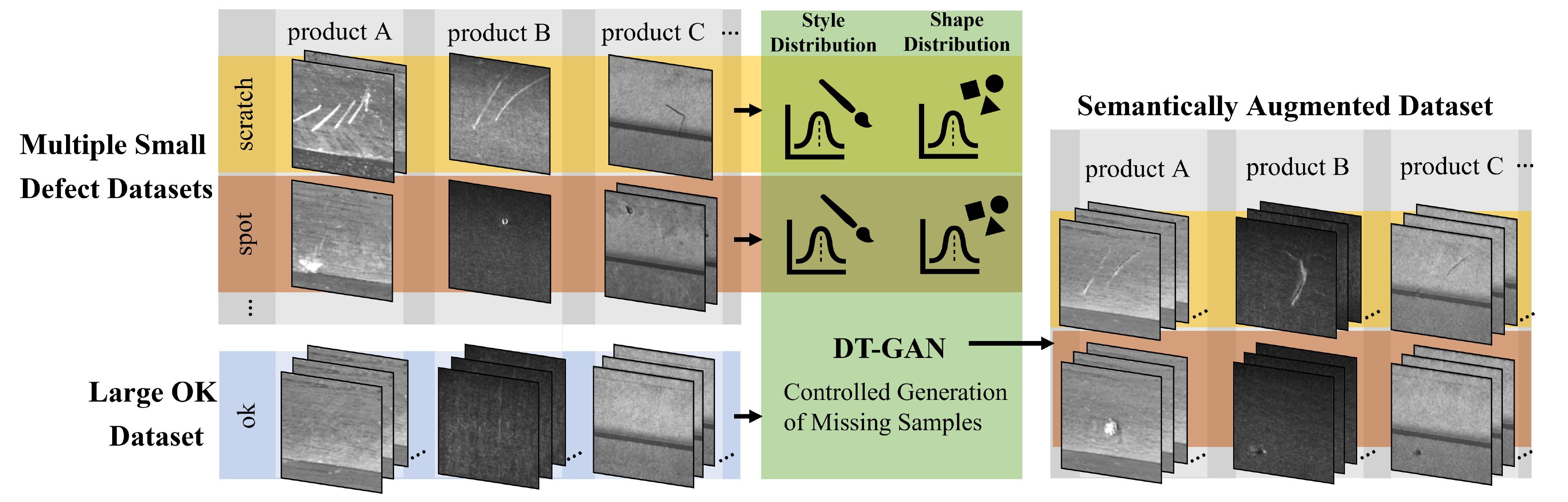}\\
\vspace{-0.3cm}
\caption{
To enrich a dataset with few defective samples, \ours{} synthesizes images under full control over background, defect shape, and style.}
\label{fig:teaser}
\end{figure*}
}

\newcommand{\figarch}{
\begin{figure*}[t]

\centering
\includegraphics[width=0.75\linewidth]{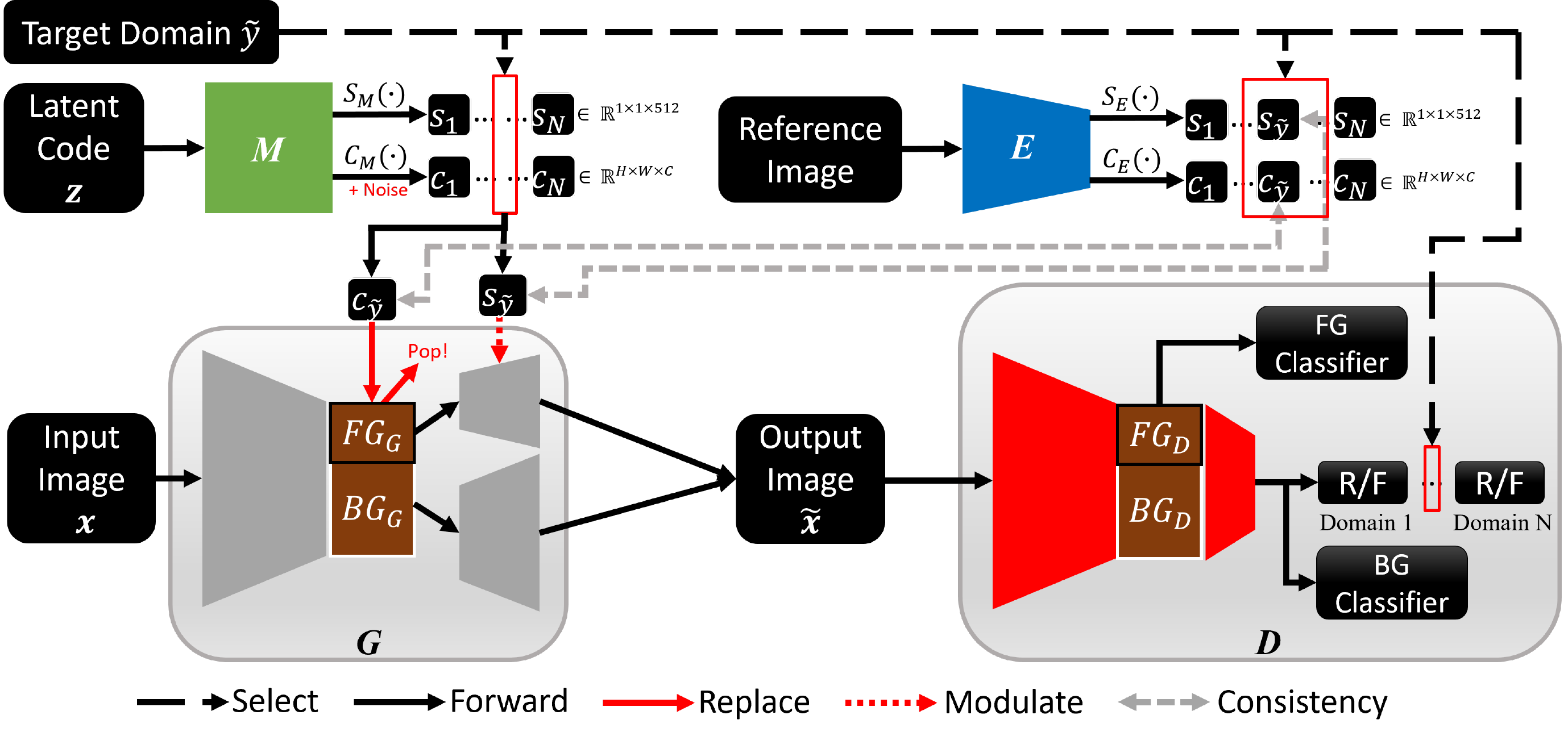}\\ 
\vspace{-0.3cm}

\caption{Overview of all modules in \ours: the mapping network $M$, the style-defect encoder $E$, the generator $G$, and the discriminator $D$. Details of the modules are in Appendix D.}
\label{fig:networks}
\end{figure*}
}

\newcommand{\figguided}{
\begin{figure*}[t]
\begin{minipage}[t]{1.0\linewidth}
\centering
\includegraphics[width=1.0\linewidth]{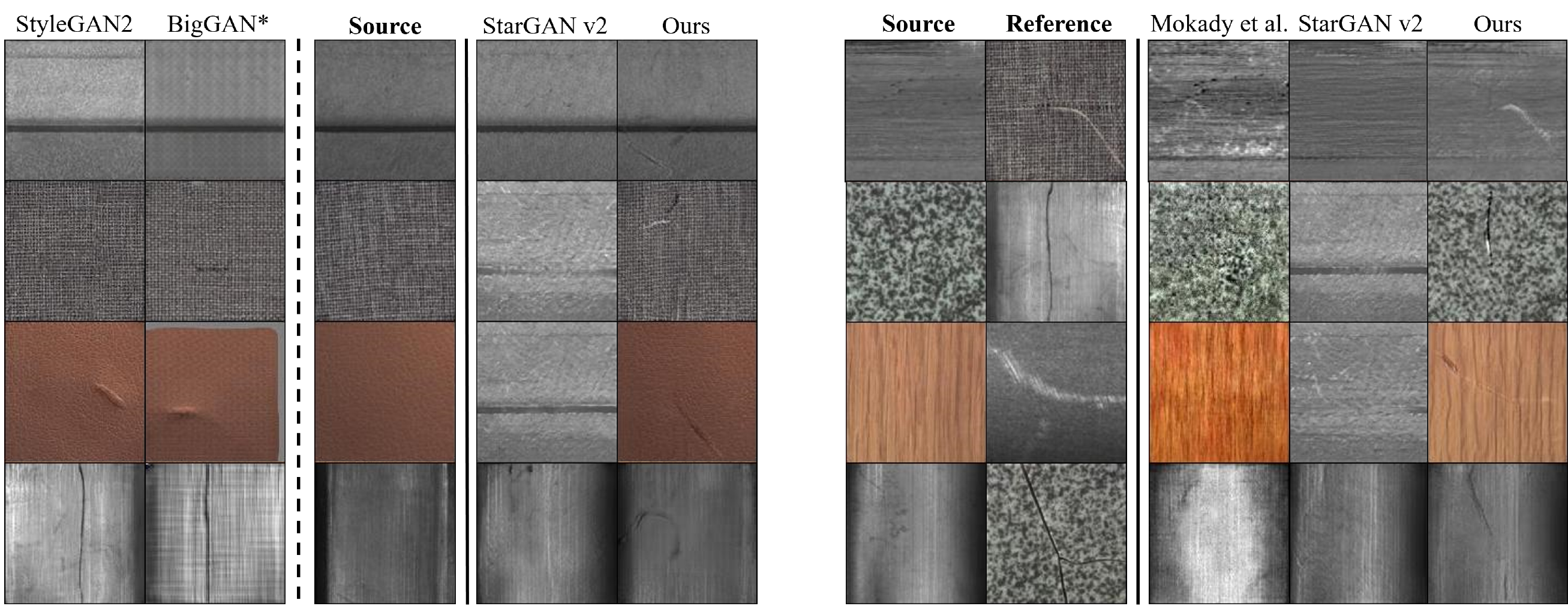}\\ %lat_guided
\makebox[0.5\linewidth][c]{\footnotesize{(a) Latent-guided}}\hfill
\makebox[0.5\linewidth][c]{\footnotesize{(b) Reference-guided}}\hfill \\
\vspace{-0.3cm}
\caption{Qualitative comparison of latent-guided and reference-guided image synthesis results on case \texttt{Normal}-to-\texttt{Scratches}. In each subfigure, the \textbf{Source} column indicates the expected background in the output images.
(a) The defective images of the first two columns are fully generated from random noise, while random defects are synthesized onto given source images in the last two columns. * indicates that the model was trained with DiffAug~\cite{zhao2020differentiable}. (b) Each method transforms the given source images into the target defect domain with the defects and styles extracted from the reference images. (Best viewed in color and zoom in.)}
\label{fig:latrefguided}
\end{minipage}
\vspace{-0.3cm}
\end{figure*}
}

\newcommand{\figab}{
\begin{figure*}[t]
\begin{minipage}[t]{1.0\linewidth}
\centering
\includegraphics[width=0.75\linewidth]{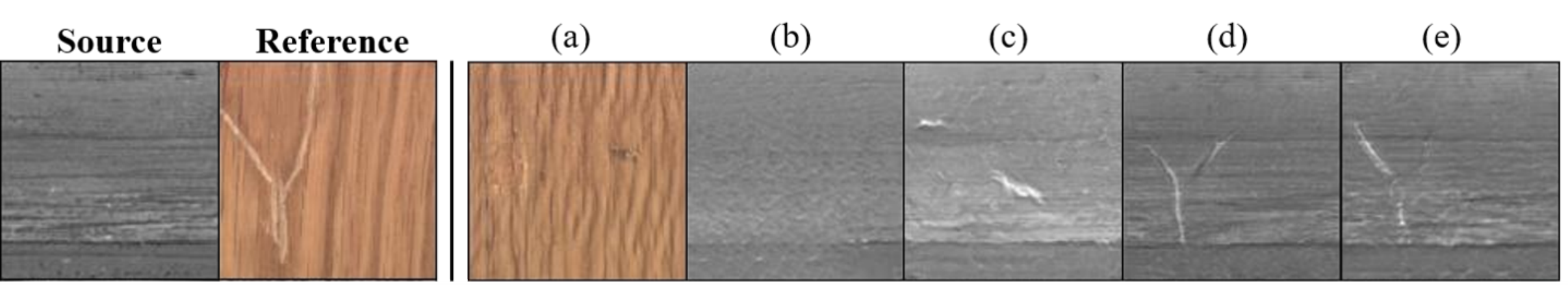}\\
\vspace{-0.3cm}
\caption{Ablation study. (a) The baseline StarGAN v2\cite{Choi_2020_CVPR}. (b) + Style-defect branches. (c) + FG and BG classifier. 
(d) + Separately decoding FG and BG in $G$. (e) + Anchor domain (e.g. \texttt{Normal}) and Noise injection in $M$. (Best viewed in color.)
}
\label{fig:ab}
\end{minipage}
\vspace{-0.1cm}
\end{figure*}
}

\newcommand{\figsty}{
\begin{figure*}[t]
\begin{minipage}[t]{1.0\linewidth}
\vspace{0.1cm}
\centering
\includegraphics[width=1.0\linewidth]{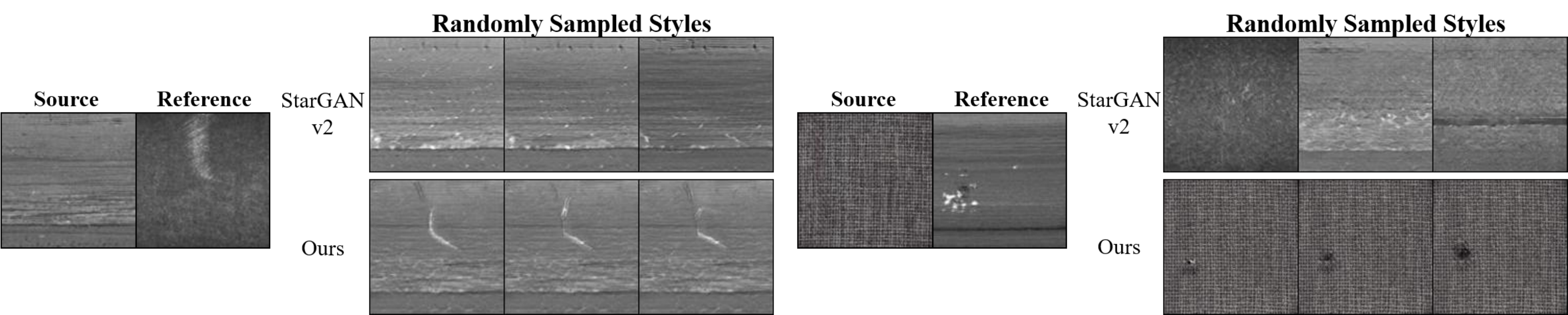}\\
\makebox[0.5\linewidth][c]{\footnotesize{(a) \texttt{Normal}-to-\texttt{Scratches}}}\hfill
\makebox[0.5\linewidth][c]{\footnotesize{(b) \texttt{Normal}-to-\texttt{Spots}}}\hfill \\
\vspace{-0.3cm}
\caption{The visual effect of randomly sampled styles on StarGAN v2 and \ours{} when given a fixed pair of source background and reference defect images.}
\label{fig:stylevary}
\end{minipage}
\end{figure*}
}

\newcommand{\tabfidkid}{
\vspace{-1.2mm}
\begin{table}[t]
\caption{Quantitative comparison of \ours{} with baseline image synthesis methods using FID(↓).  
Note that * indicates that the model is trained with augmentation methods. }
  \label{tab:kid_fid}
\centering
\newcolumntype{x}{>{\centering\arraybackslash\hspace{0pt}}p{10.5mm}}
\footnotesize{
\begin{tabular}{l@{\hspace{1.5mm}}l@{\hspace{0mm}} x@{\hspace{-0.1mm}} x@{\hspace{-0.1mm}} x@{\hspace{-0.1mm}} x@{\hspace{-0.1mm}} x@{\hspace{-0.1mm}} x@{\hspace{-0.1mm}} x@{\hspace{-0.1mm}} x@{\hspace{-0.1mm}} x }
\toprule
  &\textbf{Method} 
    & \textsc{A}           & \textsc{B}     &\textsc{C} &\rotatebox{55}{\textsc{Carpet}}  &\rotatebox{55}{\textsc{Leather}} &\rotatebox{55}{\textsc{Tile}} &\rotatebox{55}{\textsc{Wood}} &\rotatebox{55}{\textsc{MTD}} &All \\\midrule\midrule%\hline\hline %
&Mokady et al.~\cite{Mokady2020Masked}	&68.69	&66.9	&\textbf{36.21}	&41.87	&60.26	&275.12	&81.71	&68.30 &\textbf{87.38} \\
&StarGAN v2~\cite{Choi_2020_CVPR}	&96.85	&58.28	&50.95	&354.31	&336.63	&434.77	&411.37	&84.49 &228.46 \\
&StyleGAN2~\cite{Karras_2020_CVPR}	&90.1	&\textbf{52.95}	&138.09	&51.37	&\textbf{51.6}	&\textbf{225.96}	&140.01	&\textbf{51.39} &100.18 \\
&BigGAN*~\cite{Brock2019LargeSG}	&218.74	&134.41	&270.89	&34.47	&101.7	&391.54	&113.32	&67.91 &166.62 \\
&Ours	&\textbf{65.62}	&53.62	&37.94	&\textbf{27.33}	&78.01	&352.15 	&\textbf{77.11}	&78.41 &96.27
   \\\bottomrule%\hline %
\end{tabular}
}
\vspace{-1.2mm}
\end{table}
}

\newcommand{\tabqeomME}{
\vspace{-1.2mm}
\begin{table}[t]
\begin{minipage}[t]{1.0\linewidth}
\caption{Quantitative comparison of the baseline methods on the defect classification task. The reported values are the achieved error rates (\%) and standard deviation over five runs.}
  \label{tab:gen_res_all}
\centering
\newcolumntype{x}{>{\centering\arraybackslash\hspace{0pt}}p{30mm}}
\footnotesize{
\begin{tabular}{l@{\hspace{1.5mm}}l@{\hspace{3mm}} l@{\hspace{0mm}} x@{\hspace{-10mm}} x@{\hspace{-10mm}} x@{\hspace{-3mm}} }
\toprule
 &\multirow{2.5}{2em}{\textbf{Aug. Method}} &\multirow{2.5}{3em}{\textbf{Syn. Data}} &\multicolumn{3}{c}{\textbf{ResNet-50}} \\\cmidrule(lr){4-6}
 & & &{A} &{B}  &{C}\\\midrule\midrule
&None	&None	&14.91±1.52	&8.2±1.49	&15.24±1.51  \\
&Trad	&None	&13.81±2.36	&6.8±1.64	&16.57±3.20 \\
&Trad	&Mokady et al.~\cite{Mokady2020Masked}	&20.72±1.49	&5.8±2.77	&24.76±10.1  \\
&Trad	&StarGAN v2~\cite{Choi_2020_CVPR}	&10.60±1.99	&7.4±3.44	&15.81±1.44 \\
&Trad	&StyleGAN2~\cite{Karras_2020_CVPR}	&29.45±9.13	&6.8±2.05	&13.14±3.12  \\
&Trad	&BigGAN*~\cite{Brock2019LargeSG}	&12.17±1.99	&5.8±1.93	&15.62±3.06 \\
&Trad	&Ours	&\textbf{6.72±1.65}	&\textbf{4.6±0.89}	&\textbf{12.76±1.97} \\\midrule
&CutMix~\cite{Yun2019CutMixRS}	&None	&13.63±2.87	&7.4±1.52	&14.09±2.27 \\
&CutOut~\cite{Devries2017ImprovedRO}	&None	&12.36±0.50	&6.2±0.84	&12.95±2.19  \\
&MixUp~\cite{Zhang2018mixupBE}	&None	&14.36±1.75	&6.2±1.79	&16.38±2.80 \\
&CutMix~\cite{Yun2019CutMixRS}	&Ours	&14.54±3.02	&5.2±0.45	&19.42±3.47  \\
&CutOut~\cite{Devries2017ImprovedRO}	&Ours	&\textbf{12.18±1.99}	&\textbf{4.0±1.22}	&\textbf{11.42±1.50} \\
&MixUp~\cite{Zhang2018mixupBE}	&Ours	&15.27±2.98	&8.2±3.49	&21.52±3.96 
\\\bottomrule
\end{tabular}
}
\end{minipage}
\vspace{-3.2mm}
\end{table}
}

\newcommand{\tabqeomMEB}{
\vspace{-1.2mm}
\begin{table}[t]
\begin{minipage}[t]{1.0\linewidth}
\caption{Classifier performance using synthetic images generated by \ours{} trained on reduced (20A) and full-scale (All) of the SDI dataset.}
\label{tab:gen_res_syn}
\centering
\newcolumntype{x}{>{\centering\arraybackslash\hspace{0pt}}p{20mm}}
\footnotesize{
\begin{tabular}{l@{\hspace{1.5mm}}l x x x x x }
\toprule
  &\multirow{2.5}{4em}{\textbf{Dataset Size}} & \multicolumn{2}{c}{{20A}} & \multicolumn{2}{c}{{All}} \\\cmidrule(lr){3-4}\cmidrule(lr){5-6} %\multirow{2}{2em}{\textbf{}}
    &  & Trad-Aug &Ours &Trad-Aug &Ours     \\\midrule\midrule%\hline\hline%
&{A}	&34.18±4.39	&\textbf{28.55±7.32}	&13.81±2.36	&\textbf{~6.72±1.65} \\%\hline %
&{B}	&~~~5.8±0.45	&\textbf{~~~5.6±1.14}	&~~~6.8±1.64	&\textbf{~~~4.6±0.89} \\%\hline %
&{C}	&16.95±1.17	&\textbf{10.86±1.28}	&16.57±3.20	&\textbf{12.76±1.97} 
 \\\bottomrule%\hline %
\end{tabular}
}
\end{minipage}
\vspace{-1.2mm}
\end{table}
}

\newcommand{\tabqeomMEC}{
\vspace{-1.2mm}
\begin{table}[t]
\begin{minipage}[c]{1.0\linewidth}
\vspace{-0.1cm}
\caption{Cross-product defect transfer on classifiers trained with reference-guided synthetic images of \ours{} using different products as defect reference. For `v-Others', we only report the best results from all experiments with other products as reference.}

  \label{tab:cross-domain-effect-small}
\centering
\newcolumntype{x}{>{\centering\arraybackslash\hspace{0pt}}p{20mm}}
\footnotesize{
\begin{tabular}{l@{\hspace{1.5mm}}l x x x x x}
\toprule
&  & Trad-Aug & v-Same &v-Others &v-ABC     \\\midrule\midrule%\hline\hline %
 & {A}    &13.81±2.36	&11.81±2.65	&11.99±1.63 &\textbf{11.09±3.49} \\
 & {B}    &~~~6.8±1.64		&~~~6.6±1.52	&~~~6.4±1.34 &\textbf{~~~5.6±1.34} \\
 & {C}    &16.57±3.20	&14.85±1.73	&\textbf{11.23±0.80}	 &11.42±0.96
 \\\bottomrule%\hline %
\end{tabular}
}
\end{minipage}
\vspace{-1.2mm}
\end{table}
}

\newcommand{\tablpips}{
\vspace{-1.2mm}
\begin{table}[t]
\caption{Quantitative comparison of the image synthesis methods using LPIPS to measure the similarity between the synthetic samples. The lower score indicates more similarity.}
  \label{tab:lpips}
\centering
\newcolumntype{x}{>{\centering\arraybackslash\hspace{0pt}}p{10.5mm}}
\footnotesize{
\begin{tabular}{l@{\hspace{1.5mm}}l@{\hspace{0mm}} x@{\hspace{-0.1mm}} x@{\hspace{-0.1mm}} x@{\hspace{-0.1mm}} x@{\hspace{-0.1mm}} x@{\hspace{-0.1mm}} x@{\hspace{-0.1mm}} x@{\hspace{-0.1mm}} x@{\hspace{-0.1mm}} x }
\toprule
  &{\textbf{Method}} 
  & \textsc{A}           & \textsc{B}     &\textsc{C} &\rotatebox{55}{\textsc{Carpet}}  &\rotatebox{55}{\textsc{Leather}} &\rotatebox{55}{\textsc{Tile}} &\rotatebox{55}{\textsc{Wood}} &\rotatebox{55}{\textsc{MTD}}  \\\midrule\midrule%\hline\hline %
&Mokady et al.~\cite{Mokady2020Masked}	&0.34	&0.46	&0.22	&0.14	&0.28	&0.22	&0.22	&0.38  \\
&StarGAN v2~\cite{Choi_2020_CVPR}	&0.32	&0.33	&0.20	&0.37	&0.38	&0.40	&0.38	&0.38  \\
&StyleGAN2~\cite{Karras_2020_CVPR}	&0.29	&0.36	&0.19	&0.09	&0.26	&0.27	&0.18	&0.36  \\
&BigGAN*~\cite{Brock2019LargeSG}	&0.30	&0.29	&0.19	&0.08	&0.22	&0.21	&0.18	&0.37  \\
&Ours	&\textbf{0.28}	&\textbf{0.28}	&\textbf{0.17}	&\textbf{0.07}	&\textbf{0.18}	&\textbf{0.19} 	&\textbf{0.17} &\textbf{0.30} 
   \\\bottomrule%\hline %
\end{tabular}
}
\vspace{-1.2mm}
\end{table}
}

\usepackage{graphicx}
\usepackage{xspace}
\usepackage{comment}
\usepackage{amsmath,amssymb} % define this before the line numbering.
\usepackage{color}
\usepackage[accsupp]{axessibility}  % Improves PDF readability for those with disabilities.
\usepackage{hyperref}
\usepackage{url}
\usepackage{booktabs}
\usepackage{tabularx} 
\usepackage{multirow}
\usepackage{epsfig}
\usepackage{tikz}
\usetikzlibrary{tikzmark}

\usepackage[
  separate-uncertainty = true,
  multi-part-units = repeat
]{siunitx}
\usepackage{amsmath,amsfonts,bm}

\def\Secref#1{Section~\ref{#1}}
\def\eqref#1{equation~\ref{#1}}
\def\1{\bm{1}}

\DeclareMathAlphabet{\mathsfit}{\encodingdefault}{\sfdefault}{m}{sl}
\SetMathAlphabet{\mathsfit}{bold}{\encodingdefault}{\sfdefault}{bx}{n}

\newcommand{\x}{\mathbf{x}\xspace}
\newcommand{\xt}{\mathbf{\widetilde{x}}\xspace}
\newcommand{\y}{y\xspace}
\newcommand{\yt}{\widetilde{y}\xspace}
\newcommand{\ct}{\mathbf{{c}_{\yt}}\xspace}

\newcommand{\z}{\mathbf{z}\xspace}
\newcommand{\s}{\mathbf{s_{\y}}\xspace}
\newcommand{\st}{{\mathbf{{s}_{\yt}}\xspace}}
\newcommand{\cc}{\mathbf{c_{\y}}\xspace}
\newcommand{\ours}{DT-GAN}

\begin{document}
%% Enter your paper number here for the review copy
%\bmvcreviewcopy{445}

\title{Defect Transfer GAN: Diverse Defect Synthesis for Data Augmentation}

% Enter the paper's authors in order
% \addauthor{Name}{email/homepage}{INSTITUTION_CODE}
\addauthor{Ruyu Wang}{ruyu.wang@de.bosch.com}{1,2}
\addauthor{Sabria Hoppe}{sabrina.hoppe@de.bosch.com}{1}
\addauthor{Eduardo Monari}{eduardo.monari@de.bosch.com}{1}
\addauthor{Marco F. Huber}{marco.huber@ieee.org}{2,3}

% Enter the institutions
% \addinstitution{Name\\Address}
\addinstitution{
 Bosch Center for Artificial Intelligence \\
 Renningen, Germany
}
\addinstitution{
 Institute of Industrial Manufacturing and Management (IFF),\\ University of Stuttgart\\
 Stuttgart, Germany
}
\addinstitution{
 Fraunhofer IPA\\
 Stuttgart, Germany
}

\runninghead{Wang, Hoppe, Monari, Huber}{Defect Transfer GAN}

% Any macro definitions you would like to include
% These are not defined in the style file, because they don't begin
% with \bmva, so they might conflict with the user's own macros.
% The \bmvaOneDot macro adds a full stop unless there is one in the
% text already.
\def\eg{\emph{e.g}\bmvaOneDot}
\def\Eg{\emph{E.g}\bmvaOneDot}
\def\etal{\emph{et al}\bmvaOneDot}

%\newcommand{\sab}[1]{\textcolor{red}{#1}}

%-------------------------------------------------------------------------
\maketitle

\begin{abstract}
Data-hunger and data-imbalance are two major pitfalls in many deep learning approaches.
For example, on highly optimized production lines, defective samples are hardly acquired while non-defective samples come almost for free. 
The defects however often seem to resemble each other, e.g., scratches on different products may only differ in a few characteristics.
In this work, we introduce a framework, Defect Transfer GAN (DT-GAN), which learns to represent defect types independent of and across various background products and yet can apply defect-specific styles to generate realistic defective images.
An empirical study on the MVTec AD and two additional datasets showcase DT-GAN outperforms state-of-the-art image synthesis methods w.r.t. sample fidelity and diversity in defect generation.  
We further demonstrate benefits for a critical downstream task in manufacturing---defect classification. Results show that the augmented data from DT-GAN provides consistent gains even in the few samples regime and reduces the error rate up to 51\% compared to both traditional and advanced data augmentation methods.
\end{abstract}

\section{Introduction}

Automated Visual Inspection (AVI) is vital for quality control in modern production lines. 
One of the main challenges in AVI is the acquisition of suitable training data. First, labeling  is usually expensive and time-consuming. Second, only very few defective parts are produced, which leads to imbalanced datasets. Both unlabelled and imbalanced data are very challenging for neural network model training.

Generative Adversarial Networks (GANs)~\cite{goodfellow2014gan} have shown promising performance to synthesize images where real samples are lacking. However, they tend to overfit on small datasets~\cite{karras2020training}. 
In this paper, we therefore present Defect Transfer GAN (DT-GAN), 
which uses defective images across multiple products to collect more information about their shared characteristics, even for products with few defects. 
For example, a scratch-like defect on a wooden surface may share similar shapes with a scratch-like defect on a metal surface, but differ slightly in appearance according to their background materials. 
\ours{} is based on two key features: (1) a weekly-supervised
disentangling mechanism for the shared characteristics (foreground defect) 
and the unshared information (background product) 
of an input image. (2) An explicit modeling of the shape and style of foreground defects, where the styles of each defective type indicate their artistic looks such as light or heavy strokes.
Since defect-specific distributions are learned, new images can be generated with style and shape sampled randomly or extracted from reference images. 
As a result, the proposed DT-GAN achieves semantically meaningful data augmentation by  
producing novel combinations of the foreground defects, their associated style and the background products, as illustrated in Fig.~\ref{fig:teaser}. 
Moreover, by jointly modeling the defect manifold from different products that have similar defect patterns, our design not only stabilizes the GAN training but also mitigates the overfitting issue on limited data.

Experiments on three industrial-oriented datasets showcase the power of DT-GAN in both defect synthesis and its usefulness in a downstream defect classification task where we report up to 51\% reduction in error rates when augmenting the data with DT-GAN.
\figteaser

\section{Related Work}

Surface defect inspection 
aims at identifying and classifying defects with the help of machine vision. Traditional methods~\cite{10.1016/j.imavis.2011.02.002,  10.1016/j.neucom.2013.07.038}
build models upon hand-crafted feature extractors, which are 
often outperformed by deep learning based models. However, the performance and generalization ability of deep learning approaches are restricted due to a limited number of defective samples in real-world scenarios. 

Insufficient data has been addressed by multiple methods.
Among them, data augmentation aims to enrich the training dataset by introducing
invariances for the model to capture. Apart from the traditional augmentations~\cite{Perez2017TheEO, Shorten2019ASO} such as random flipping and cropping, 
some more advanced regularization techniques~\cite{Devries2017ImprovedRO,Zhang2018mixupBE} like
Cutmix~\cite{Yun2019CutMixRS} have been proposed. 
However, they do not introduce semantically new information to the training set.

In contrast, GANs augment data with meaningful semantic transformations.
The power of GANs has been demonstrated in many computer vision tasks such as image synthesis~\cite{Brock2019LargeSG, pmlr-v97-lucic19a, donahue2019bigbigan}, image to image translation~\cite{Isola2017ImagetoImageTW, Zhu2017TowardMI, Huang2018MultimodalUI, ma2018exemplar, park2019spade, Mokady2020Masked}, style translation~\cite{7780634, johnson2016perceptual, Choi_2018_CVPR}, image impainting~\cite{Yeh2017SemanticII, 8578675,
Pathak2016ContextEF, Yu2019FreeFormII} and many other applications. 
Several recent works~\cite{9000806, Zhang_2021_WACV} have proposed to use GANs for data augmentation with realistic defective samples. 
Defect-GAN~\cite{Zhang_2021_WACV} for instance tries to capture the stochastic variation within defects by mimicking the defacement and restoration processes.
However, it still learns a deterministic mapping between inputs and outputs while our \ours{} achieves multi-modality by varying styles. Moreover, \ours{}
incorporates the shared characteristics of defects from multiple products, which further enrich the diversity of synthetic defects for each product.

\section{Methodology}
\label{sec:methodology}

Our approach is cast as an unpaired image-to-image translation problem, where we aim to achieve domain transfer between multiple domains within a single model. 
We define the \emph{domain} as foreground defect types, where each type of defect is associated with a style distribution describing the artistic looks. The background product of an input should remain unaffected during the translation.

\subsection{Proposed Framework}
\label{sec:method}

Our framework builds on StarGAN v2~\cite{Choi_2020_CVPR}, which transforms an image by a single vector representing the target style for the full image. However, to generate semantically meaningful defective images in our setting, it is essential for the model to understand and allow control over the components in an input image---the foreground defect pattern with its associated style and the background product.
Given an image $\x \in {\cal X}$, its original defect domain $y \in {\cal Y}$ and its background product $p \in {\cal P}$, we modify and extend all four modules from~\cite{Choi_2020_CVPR} as follows (see Fig.~\ref{fig:networks} for the resulting model).

\noindent\textbf{Style-Defect Separation.} 
Our method models the shape and style separately by a domain-specific defect $\cc \in\mathbb{R}^{H\times W \times C}$ and a style vector $\s \in\mathbb{R}^{1\times 1 \times 512}$. The former learns to capture the shapes of the defects and the latter models their artistic looks. This feature allows our method to produce non-deterministic outputs by varying the style when the same target defect ($\ct$) is given.
Thus, the mapping network $M$ is trained to generate both defect patterns and their styles in all domains from a latent code $\z$. The final outputs  $(\ct,\st) = {M}_{\yt}(\z)$ are selected by the given target domain $\yt$ among $N$ output branches.
The procedure for the style-defect encoder $E$ is similar, except that the domain-specific defect $\ct$ and style $\st$ are extracted from a given reference image. 
%Note that the reference image can be any image including the generated one. 
The two subnetworks are coupled by a consistency constraint (discussed in \Secref{sec:traning-objectives}) between the joint image-style spaces, which prevents model degeneration and retains the multi-modality.

\noindent\textbf{Foreground/Background ($\mathbf{FG}$/$\mathbf{BG}$) Disentanglement.} 
It is crucial to identify and disentangle the FG and BG of an input image for achieving control over the defect (i.e., FG) and retraining the BG. The FG/BG disentanglement is performed by a depth-wise split at the bottleneck of $G$, which divides the feature map into two parts.
Driven by the classification losses as discussed in \Secref{sec:traning-objectives}, the model encodes the BG into the first channels and the domain-specific defect $\cc$ into the latter channels. Instead of translating via a style vector, \ours{} achieves domain transfer by altering the feature map---where $\cc$ is replaced with defect $\ct$ from the target domain.
The given defect style $\st$ is applied through the adaptive instance normalization (AdaIN)~\cite{Huang_2017_ICCV}. However, to modulate only the fine details of the target defect $\ct$, the background $BG_G$ is decoded separately without style modulation. 
Finally, $BG_G$ and $\ct$ are concatenated together by depth-wise pooling before output.
This design breaks the conditional relationship between FG and BG and therefore enables our method to freely combine them as well as learn the full variation of foreground defects.

\noindent\textbf{Multi-Task Discriminator with Auxiliary Classifiers.} 
An FG defect classifier and a BG classifier are deployed in the multi-task discriminator  to strengthen the disentanglement of FG and BG.
Independent of the background, the FG defect classifier identifies the specific defect from an input image $\x$ in the latent space. The BG classifier acts on the image-level and decides whether the background information of the input image is well preserved.
Apart from that, each branch ${D}_{\yt}$ in the multi-task discriminator $D$ is trained to determine if an image $\x$ is a real image of its foreground defect domain or a fake image $\xt$ generated by $G$. 

\noindent\textbf{Anchor Domain and Noise Injection.} 
We impose an additional constraint on the generated and extracted foreground defect of a normal sample by setting the latent representation $\ct$ to zero~\cite{Mokady2020Masked}. We refer to this constraint as the `anchor domain' and hypothesize that it supports the FG/BG disentanglement. Moreover, inspired by~\cite{Karras_2019_CVPR}, a per-pixel noise injection is introduced to $M$ to improve the diversity of the generated defects.

\figarch
\subsection{Training Objectives}
\label{sec:traning-objectives}

\noindent\textbf{Adversarial Loss.} 
We follow the same adversarial loss as in~\cite{Choi_2020_CVPR} to encourage an output image $\xt = G(\x, \ct, \st)$ to be indistinguishable from real images in the target domain $\yt$
\begin{equation}
\begin{split}
 \mathcal{L}_{\text{adv}} =  \thinspace \mathbb{E}_{\x, \y} \big[ \log{{D}_{\y}(\x)} \big]   +  \thinspace \mathbb{E}_{\x, \yt, \z}[\log{(1 - {D}_{\yt}(\xt))}]~,
\end{split}
\label{eqn:adv_loss}
\end{equation}
\noindent where ${D}_{y}$ and ${D}_{\yt}$ are the output branches of $D$ that correspond to the source domain $y$ and the target domain $\yt$, respectively.

\noindent\textbf{Style-Defect Reconstruction Losses.} 
To ensure 
$G$ takes the domain-specific defect $\ct$ and the style $\st$ into consideration during the generation process, we employ a style-defect reconstruction loss (cf. the gray dashed arrows in Fig.~\ref{fig:networks})
\begin{equation}
\begin{split}
 \mathcal{L}_{\text{sd\_rec}} &=  \mathbb{E}_{\x, \yt, \z} \big[{\lVert \ct - C_E(\xt)\rVert}_{1} \big]  \thinspace +  \mathbb{E}_{\x, \yt, \z} \big[{\lVert \st - S_E(\xt) \rVert}_{1} \big]~,
\end{split}
\label{eqn:sty_rec_loss}
\end{equation}
\noindent where $C_E(\cdot)$ and $S_E(\cdot)$ indicate the extracted defect and style of an input, respectively.
This objective urges %the style-defect encoder 
$E$ to recover $\ct$ and $\st$ from $\xt$. 
Besides, we apply another constraint to enforce that the detached domain-specific defect from $G$ is consistent with the one retrieved from $E$ 
\begin{equation}
\begin{split}
\mathcal{L}_{\text{d\_rec}} = \mathbb{E}_{\x, \y, \yt, \z} \big[ {\lVert FG_G(\x) - \cc \rVert}_{1} \big] + \mathbb{E}_{\x, \y, \yt, \z} \big[ {\lVert FG_G(\xt) - \ct \rVert}_{1} \big]~,
\end{split}
\end{equation}
\noindent where $\cc = {E}_{y}(\x)$, $\ct = {E}_{\yt}(\xt)$; $FG_G(\x)$ and $FG_G(\xt)$ are the replaced defect from input image $\x$ and generated image $\xt$, respectively.

\noindent\textbf{Diversity Loss.} 
For a pair of random latent codes $\z_1$ and $\z_2$, we compute 
 $\ct_i, \st_i = {M}_{\yt}(\z_i)$ for $ i \in \{1,2\}$
 and enforce a different outcome of $G$ for differently mixed defect and style input pairs according to
\begin{equation}
\begin{split}
 \mathcal{L}_{\text{ds}} &= \mathbb{E}_{\x, \yt, \z_1, \z_2} \big[ { \lVert G(\x,\ct_1, \st_2) - G(\x,\ct_2, \st_1) \lVert}_{1} \big]\\ & \thinspace + \mathbb{E}_{\x, \yt, \z_1, \z_2} \big[ { \lVert G(\x,\ct_1, \st_1) - G(\x,\ct_2, \st_2) \lVert}_{1} \big] \\
 & \thinspace + {\Sigma}_{m,n,o}\big[ \mathbb{E}_{\x, \yt, \z_1, \z_2} \big[ { \lVert G(\x,\ct_m, \st_n) - G(\x,\ct_o, \st_o) \lVert}_{1}   \big]\big]~,
\end{split}
\label{eqn:ds_loss}
\end{equation}
\noindent where $m, n \in \{1,2 | m \neq n\}$ and $o \in \{1,2\}$. Driven by this term, 
$G$ is forced to discover meaningful defects and style features that 
lead to diversity in generated images.
We ignore the denominator ${\lVert \z_1 - \z_2 \lVert}_{1}$ of the original diversity loss~\cite{Mao2019ModeSG} for stable training as in~\cite{Choi_2020_CVPR}.

\noindent\textbf{Cycle Consistency Loss.}
To encourage the disentanglement of the background, the domain-specific defect and the style, we impose a cycle consistency loss~\cite{Zhu_2017_ICCV} to reconstruct the input image $\x$ with given defect $\cc$ and style $\s$
\begin{equation}
\begin{split}
\mathcal{L}_{\text{cyc}} &= \mathbb{E}_{\x, \y, \yt, \z} \big[ {||\x - G(\xt, \cc, \s)||}_{1} \big]~,
\end{split}
\end{equation}
\noindent where $\cc, \s = {E}_{y}(\x)$ is the
defect and style of the input image $\x$, respectively. 

\noindent\textbf{Classification Losses.} We employ two classification losses, which are essential to enforce the FG/BG disentanglement: 
First, the FG defect classification loss
\begin{equation}
\begin{split}
 \mathcal{L}_{\text{FG}} = \mathbb{E}_{{\x}_{\text{real}}, y} \big[ -\log{{D}_\text{FG}(y|{\x}_{\text{real}})}  \big] + \mathbb{E}_{{\x}_{\text{fake}}, \yt} \big[ -\log{{D}_\text{FG}(\yt|{\x}_{\text{fake}})}  \big]~,
\end{split}
\end{equation}
\noindent which aims to ensure %that 
the domain-specific defect is properly encoded and carries enough information from the target domain. Second, the BG classification loss
\begin{equation}
\begin{split}
 \mathcal{L}_{\text{BG}} = \mathbb{E}_{{\x}_{\text{real}}, p} \big[ -\log{{D}_\text{BG}(p|{\x}_{\text{real}})}  \big] + \mathbb{E}_{{\x}_{\text{fake}}, p} \big[ -\log{{D}_\text{BG}(p|{\x}_{\text{fake}})}  \big]~,
\end{split}
\end{equation}
\noindent where $p$ is the corresponding background type of ${\x}_{\text{real}}$ and ${\x}_{\text{fake}}$. With the help of this objective, $G$ learns to preserve the unshared characteristics of its input image $\x$ while dissociating the foreground defect.

We summarize the full objective and provide the training details in Appendix B. 

\section{Experiments}

\noindent\textbf{Dataset.}
We conducted the image synthesis experiments on three industrial-oriented datasets: the MVTec AD, the Magnetic Tile Defects (MTD), and a new dataset of industrial images---the Surface Defect Inspection (SDI)\footnote{The SDI dataset will be published with the final version of the paper.}. For all the experiments, we re-organized the defects in the datasets into three mutually exclusive classes: \texttt{Normal}, \texttt{Scratches}-like and \texttt{Spots}-like according to their visual appearance. 

All three datasets are relatively small---the number of defective images for each defect category varies from 8 to 620, which is rather limited considering the sophisticated patterns of defects. This poses a major challenge for training generative models. Details of the datasets are summarized in Appendix A.1.

To study the performance of DT-GAN generated samples in defect classification, all the experiments were performed on the SDI dataset due to the limited availability of defective samples in the other two datasets. Note that only the training set of the SDI datset was used in GAN training, the validation and test set were left untouched for final evaluation in classifier training. For a fair comparison, all images were resized to $128 \times 128$ resolution for both GAN training and classifier training, which was also the highest resolution used in the baselines for image generation.

\subsection{Defect Generation}

\label{sec:defect-generation-baselines}
\noindent\textbf{Baselines.} 
As discussed in \Secref{sec:methodology}, \ours{} can either use $M$  
to randomly generate defects and styles, or use $E$
to extract both from one or two reference images. 
We refer to these cases as `latent-guided' and `reference-guided', respectively.
Since the two ways of guidance are fundamentally different, we evaluated them against two sets of baselines:
our reference-guided image generation was compared to Mokady et al.~\cite{Mokady2020Masked} and StarGAN v2~\cite{Choi_2020_CVPR}. 
Note that without the key designs we introduced in Section~\ref{sec:method}, \ours{} degrades to~\cite{Choi_2020_CVPR}. 
Images generated through the latent-guided part of \ours{} were compared to the state-of-the-art GANs in image synthesis:
BigGAN~\cite{Brock2019LargeSG} and StyleGAN2~\cite{Karras_2020_CVPR}. We set both~\cite{Brock2019LargeSG} and~\cite{Karras_2020_CVPR}\footnote{We used the implementation in~\cite{karras2020training} for conditional training.} to condition on defect types during training.
All baselines were trained from scratch with the public implementations provided by the authors\footnote{ We could not obtain  the code of Defect-GAN~\cite{Zhang_2021_WACV} to reproduce their results.}.

\noindent\textbf{Metrics.}
We employed the commonly used frechet inception distance (FID)~\cite{Heusel2017GANsTB} to evaluate both the visual quality and the diversity of the generated images. 
A lower FID score indicates better performance.

\figab
\smallskip
\noindent\textbf{Ablation Study.}
We visually demonstrate the effect of each feature we added to \ours{} compared to~\cite{Choi_2020_CVPR} in Fig.~\ref{fig:ab}, using the examples of reference-guided image synthesis from \texttt{Normal} to \texttt{Scratches}. Also, we report the average FID over all three datasets for each configuration in Appendix E.1.

Fig.~\ref{fig:ab}(a) corresponds to~\cite{Choi_2020_CVPR} and highlights the drawback of an entangled style vector---the model extracts a style from the entire reference image instead of a style of the foreground defect and thus, changes the background product in its output, which we refer to as an `identity-shift'.
We first tackle this problem by modeling the foreground defect and style explicitly and introducing the FG/BG disentanglement, so the defect replacement and style modulation can be performed at different stages in $G$. 
This leads to better preservation of the background structure in (b), but the resulting image contains no clear defect from the reference image. 
Thus, we add a FG and a BG classifier to $D$ 
in (c) to ensure the output image contains the desired foreground defect.
Note that the additional product type labels can be acquired automatically from production lines. These two auxiliary classifiers improve the image quality by a big margin, however, the generated defects fail to preserve the structure shown in the given reference. 
To address this issue, we add separate decoders for FG and BG in $G$.
As seen in Fig.~\ref{fig:ab} (d), this enhances the preservation of background characteristics like lighting even more and the foreground defect characteristics start to match the patterns from the reference.
Finally, we impose the anchor domain constraint 
and the per-pixel noise injection to $M$. 
This leads to more diverse defects which are not clear in Fig.~\ref{fig:ab} (e) but clearly affect the FID scores.

\tabfidkid
\smallskip
\noindent\textbf{Quantitative and Qualitative Evaluation.}
\label{sec:quantitative-evaluation}
The quantitative comparison of \ours{} with baseline image synthesis methods on all datasets is shown in \autoref{tab:kid_fid}, and the qualitative comparison is in Fig.~\ref{fig:latrefguided}. For a fair comparison, we trained~\cite{Brock2019LargeSG},~\cite{Karras_2020_CVPR} and~\cite{Mokady2020Masked} on each product separately to have control on background products. We also experimented with augmentation methods for GAN training~\cite{karras2020training, zhao2020differentiable} and only report the best setting (see Appendix E.4).
Note however, the images from~\cite{Choi_2020_CVPR} and \ours{} were always obtained from a single model. 

As observed in \autoref{tab:kid_fid}, our method outperforms the rest in 3 out of 8 cases and provides the second best overall performance. Our method is often outperformed by \cite{Karras_2020_CVPR} and~\cite{Mokady2020Masked}, however, we note that FID is not sensitive to detect overfitting, which often happens when training on a small dataset. We thus present the nearest neighbor results in Appendix E.5 to demonstrate that the low FID scores of~\cite{Karras_2020_CVPR} and~\cite{Mokady2020Masked} come from memorizing the training dataset. Also, there are further evaluations in the downstream task which support our assumption (see Appendix C.2).

The latent-guided image synthesis results are presented in Fig.~\ref{fig:latrefguided} (a). We observe that generated samples from~\cite{Brock2019LargeSG} often present abnormal grid patterns and samples from~\cite{Karras_2020_CVPR} either overfit %the training data 
or contain no clear defect.
Both methods do not take images as inputs but infer both FG and BG of a synthetic image from a given latent code.  
This conditioning leads to limited diversity and artifacts in the output images while making the models less robust to overfitting. 
In contrast,~\cite{Choi_2020_CVPR} performs translation based on input images but suffers from the same entanglement issue. 
As discussed in the ablation study, due to the lack of our designed features,~\cite{Choi_2020_CVPR} fails to preserve the product type in its output images (i.e., identity-shift) and results in undesired outputs, which is also reflected in the FID scores.
Our architecture, which disentangles FG and BG, mitigates these issues and provides visually convincing results.

Also for reference-guided image synthesis, where we used defects from different foreground reference images as illustrated in Fig.~\ref{fig:latrefguided} (b),
only our method produces high-quality images with preserved background from the source and transferred foreground defect from the reference. This again showcases the effectiveness of the FG/BG disentanglement.
See Appendix E.2 for more images, where \ours{} is the only method that can perform translations between all domains, including defect-to-defect translations.

\figguided

\figsty
\noindent\textbf{Styling.}
We visually demonstrate the effect of the style vector in Fig.~\ref{fig:stylevary}. 
When combing randomly sampled styles with a fixed pair of input images,~\cite{Choi_2020_CVPR} suffers from the identity-shift and fails to produce meaningful defects while \ours{} provides a variety of artistic styles in its outputs due to the style-defect separation as discussed in Section~\ref{sec:method}.

\tabqeomME
\tablpips
\tabqeomMEB
\tabqeomMEC

\subsection{\ours{} for Data Augmentation}
\label{sec:ours-for-da}
To demonstrate the effectiveness of our synthetic images, we also evaluated \ours{} as a data augmentation method for defect classification as an exemplary downstream task. Therefore, we used all defective samples from the SDI dataset and an additional 4,000 normal (non-defective) images for each product to generate further defective samples for classifier training (see Appendix C for more details). 

As backbone we employed the widely used ResNet-50~\cite{He2016DeepRL} with ImageNet pretrained weights. 
For experiments with synthetic data, we attached an auxiliary classifier to the network through a Gradient Reversal Layer (GRL)~\cite{pmlr-v37-ganin15}, ensuring the extracted features by the backbone are invariant for both the real and the synthetic samples.
Since the SDI dataset is highly imbalanced, we oversampled the minority classes~\cite{And98datamining} unless the data was balanced through synthetic images.
Traditional data augmentations like random horizontal flips, jittering, and lighting~\cite{Shorten2019ASO} were always applied unless otherwise specified. 
All following results were evaluated by the achieved error rates over five runs with different random seeds.

\noindent\textbf{Effectiveness of Synthetic Data.} 
We first compare the classifier performance for no augmentation, 
traditional data augmentation (Trad-Aug), advanced regularization techniques like \cite{Yun2019CutMixRS}, \cite{Devries2017ImprovedRO} and \cite{Zhang2018mixupBE} and a combination of traditional augmentation with synthetic images for GAN methods including \ours{} in \autoref{tab:gen_res_all}. For brevity, the detailed results are in Appendix E.3. We found consistent improvements when combining  our method with \cite{Devries2017ImprovedRO}. In contrast, \cite{Yun2019CutMixRS} and \cite{Zhang2018mixupBE} seemed to jeopardize the performance. We hypothesize that it is due to the real-fake domain gap---both methods regularize the training by randomly concatenating two training images in different manners, however, it destructs the backpropagation from the GRL, which results in poor performance.

Methods like \cite{Mokady2020Masked} and \cite{Karras_2020_CVPR} outperformed our method in some cases concerning FID. However, our method led to better performance in downstream classifier training. We hypothesize this is because the baselines overfit the training set. 
Quantitatively this is supported by the LPIPS \cite{zhang2018perceptual} scores in \autoref{tab:lpips}, which computes the similarity of the synthetic samples to each other. 
Qualitatively this could be seen in the nearest neighbor analysis in Appendix E.5. None of the commonly used metrics is designed to indicate small perturbations like the variance of defects. However, by combining FID, LPIPS, the nearest neighbor analysis, and the classifier performance, we believe that our method improves performance on all products due to the combination of high visual  quality and diversity in our samples.

\noindent\textbf{Impact of Dataset Size.}
Motivated by the limited availability of data in real-world production scenarios, 
we evaluated \ours{} for data augmentation on the full SDI dataset (All) as well as a subset, which only contains 20 defective samples of product A for each defect type (20A). In this case, \ours{} was also trained on the reduced subset.
As shown in \autoref{tab:gen_res_syn}, there is a clear improvement when synthetic images from \ours{} are used as data augmentation, even for the extremely limited data subset.

\noindent\textbf{Cross-Product Defect Transfer.}
We hypothesized that limited data can be counteracted by transferring defects across multiple background products if there are at least some defects that occur on multiple products.
We tested this approach by comparing the performance of classifiers trained on synthetic images with defects from the same product (v-Same), 
from another product (v-Others) and from all products (v-ABC).
As we can see in \autoref{tab:cross-domain-effect-small}, the best performances are achieved by the models that transfer defects across products (v-Others or v-ABC). We interpret this as support for our hypothesis and its practical usefulness.
The full scale results are in Appendix E.3.

\section{Conclusion}
We propose a novel method, \ours,
which allows diverse defect synthesis and semantic data augmentation by exploiting shared defect characteristics across multiple products.
Due to explicit style-defect separation and FG/BG disentanglement, DT-GAN achieves higher image fidelity, better variance in defects, and full control over FG and BG while being sample-efficient and robust against model overfitting.
We demonstrated the feasibility and benefits of \ours{} on a real industrial defect classification task and the results show that our method provides consistent gains even with limited data and boosts the performance of classifiers up to 51\% compared to traditional augmentation and state-of-the-art image synthesis methods.
For future investigation, we aim to represent defects and their styles more explicitly (e.g., localization), improve the explainability of the model and also enhance the model transferability to unseen products.

\clearpage
% ---- Bibliography ----
%\include{bibcommand}
%\clearpage
%\include{appendix}

\end{document}